\documentclass[runningheads]{llncs}
\usepackage{graphicx}

\usepackage{tikz}
\usepackage{comment}
\usepackage{amsmath,amssymb} 
\usepackage{color}

\usepackage[accsupp]{axessibility}  

\usepackage{epsfig}
\usepackage{graphicx}
\usepackage{graphicx}
\usepackage{multirow}
\usepackage{enumitem}
\usepackage{mathrsfs}
\usepackage{booktabs}  
\usepackage{threeparttable}
\usepackage{setspace}
\usepackage{amsmath}
\usepackage{url}
\usepackage{colortbl}
\usepackage{rotating}
\usepackage{makecell}
\usepackage{diagbox}
\usepackage{array}
\usepackage{algorithmic}
\usepackage{algorithm}
\usepackage{tabularx}
\newcommand{\PreserveBackslash}[1]{\let\temp=\\#1\let\\=\temp}
\newcolumntype{C}[1]{>{\PreserveBackslash\centering}p{#1}}
\newcolumntype{R}[1]{>{\PreserveBackslash\raggedleft}p{#1}}
\newcolumntype{L}[1]{>{\PreserveBackslash\raggedright}p{#1}}

\usepackage{marvosym}
\makeatletter
\newcommand{\printfnsymbol}[1]{%
	\textsuperscript{\@fnsymbol{#1}}%
}
\makeatother

\usepackage[pagebackref,breaklinks,colorlinks]{hyperref}

\begin{document}
\pagestyle{headings}
\mainmatter
\def\ECCVSubNumber{4873}  

\title{Towards Real-World HDRTV Reconstruction: \\ A Data Synthesis-based Approach} 


\titlerunning{Towards Real-World HDRTV Reconstruction}
%
\author{Zhen Cheng\inst{1}\thanks{Equal contribution. This work was done when Zhen Cheng was an intern in Huawei Noah's Ark Lab.}\and
Tao Wang\inst{2}\printfnsymbol{1} \and
Yong Li\inst{2} \and
Fenglong Song\inst{2} \and
Chang Chen\inst{2} \and
Zhiwei Xiong\inst{1}\textsuperscript{\Letter}}
\authorrunning{Z. Cheng et al.}
%
\institute{University of Science and Technology of China \\
	\email{mywander@mail.ustc.edu.cn,zwxiong@ustc.edu.cn} \and
Huawei Noah's Ark Lab \\
\email{\{wangtao10,liyong156,songfenglong,chenchang25\}@huawei.com}}
\maketitle

\begin{abstract}
Existing deep learning based HDRTV reconstruction methods assume one kind of tone mapping operators (TMOs) as the degradation procedure to synthesize SDRTV-HDRTV pairs for supervised training. 
In this paper, we argue that, although traditional TMOs exploit efficient dynamic range compression priors, they have several drawbacks on modeling the realistic degradation: information over-preservation, color bias and possible artifacts, making the trained reconstruction networks hard to generalize well to real-world cases. 
To solve this problem, we propose a learning-based data synthesis approach to learn the properties of real-world SDRTVs by integrating several tone mapping priors into both network structures and loss functions. 
In specific, we design a conditioned two-stream network with prior tone mapping results as a guidance to synthesize SDRTVs by both global and local transformations.
To train the data synthesis network, we form a novel self-supervised content loss to constraint different aspects of the synthesized SDRTVs at regions with different brightness distributions and an adversarial loss to emphasize the details to be more realistic.
To validate the effectiveness of our approach, we synthesize SDRTV-HDRTV pairs with our method and use them to train several HDRTV reconstruction networks.
Then we collect two inference datasets containing both labeled and unlabeled real-world SDRTVs, respectively. 
Experimental results demonstrate that, the networks trained with our synthesized data generalize significantly better to these two real-world datasets than existing solutions.

\end{abstract}

\section{Introduction}

Recent years have seen the huge progress on ultra high-definition (UHD) display devices such as OLED~\cite{geffroy2006organic}, which can display high dynamic range television sources (HDRTVs) with high dynamic range (HDR, \emph{e.g.}, 10 bit quantization) and wide color gamut (WCG, \emph{e.g.}, BT.2020~\cite{bt2020}). However, while such HDR display devices (named HDR-TVs) become more popular, most available images/videos are still standard dynamic range television sources (SDRTVs). 

\begin{figure*}[!t]
	\centering
	\includegraphics[width=0.88\linewidth]{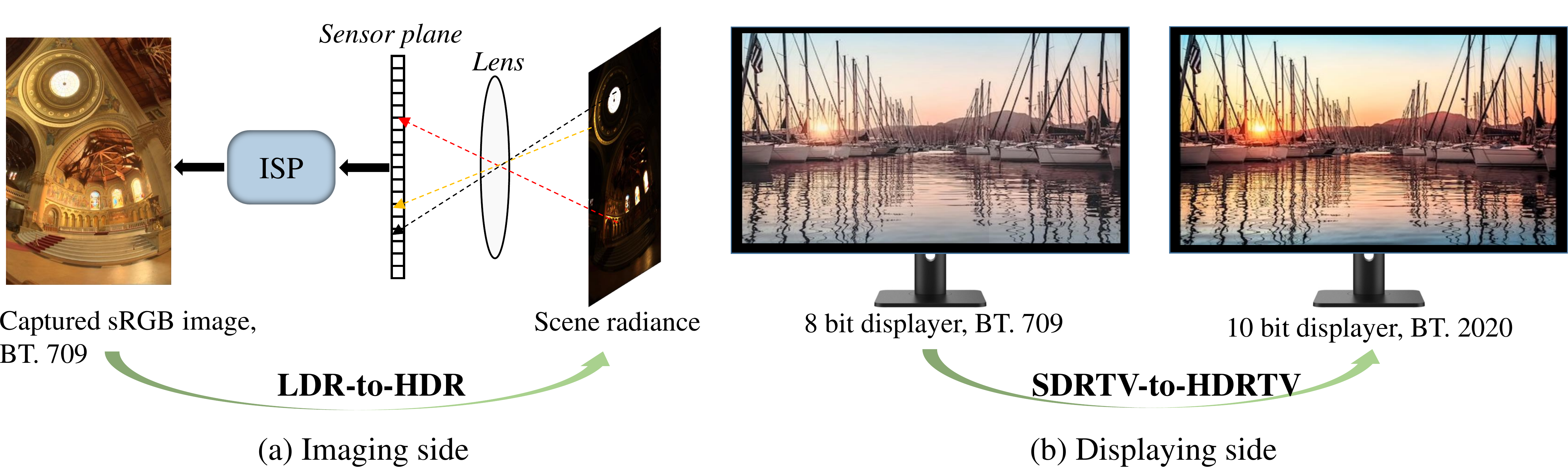}
	\caption{Illustration of the difference between the tasks LDR-to-HDR (at the imaging side) and SDRTV-to-HDRTV (at the displaying side).}
	\label{Fig_imaging_display}
\end{figure*}

To this end, previous researches~\cite{banterle2006inverse,rempel2007ldr2hdr,kovaleski2009high,10.1145/3130800.3130834,eilertsen2017hdr,lee2018deep,liu2020single,santos2020single} focus on recovering the linear and scene radiance maps from the captured sRGB sources, forming the LDR-to-HDR problem defined at the imaging side, as shown in Fig.~\ref{Fig_imaging_display}(a). Then the scene radiance maps are transformed to HDRTVs via complicated post-processing~\cite{kim2019deep,kim2020jsi,chen2021new}. However, such post-processing has been not well-defined for the standards of HDRTVs, resulting in severe color bias and artifacts~\cite{kim2020jsi,chen2021new}. Recently, researchers introduced deep learning techniques to straightforwardly reconstruct HDRTVs from their corresponding SDRTVs~\cite{kim2019deep,zeng2020sr,kim2020jsi,chen2021new}, forming the problem SDRTV-to-HDRTV at the dispalying side (Fig.~\ref{Fig_imaging_display}(b)). Such solutions need to train convolutional neural networks (CNNs) relying on SDRTV-HDRTV pairs. Hence, the acquisition of such paired data becomes a vital problem.

There exists two possible ways to get SDRTV-HDRTV pairs: acquisition by cameras and synthesis by algorithms. The former acquires SDRTV-HDRTV pairs via asynchronous camera shots like those in super-resolution~\cite{chen2019camera,cai2019toward}. However, such approach faces difficulties to get large datasets for network training due to its high sensitivity to motion and light condition changes. The latter solution can be further divided into two categories: camera pipeline based and tone mapping operator (TMO) based. Camera pipeline based approaches get the scene radiance map first and then process it to SDRTV and HDRTV via different processing pipelines. However, mostly the processing from light radiance to HDRTV is unknown, which makes the solution unavailable~\cite{chen2021new}. In consequence, existing SDRTV-to-HDRTV methods rely on TMOs~\cite{tmo_liang,tmo_chiu,tmo_hable,tmo_kuang,tmo_raman,tmo_reinhard,tmo_wardHistAdj} that compress the dynamic range via global or local transformations as the degradation procedure to synthesize the SDRTV data.

\begin{figure*}[!t]
	\begin{center}
		\begin{minipage}{0.48\linewidth}
			\begin{minipage}{0.32\linewidth}
				\centerline{\includegraphics[width=1\linewidth]{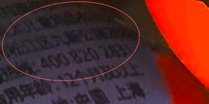}}
			\end{minipage}
			\hfill
			\begin{minipage}{0.32\linewidth}
				\centerline{\includegraphics[width=1\linewidth]{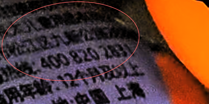}}
			\end{minipage}
			\hfill
			\begin{minipage}{0.32\linewidth}
				\centerline{\includegraphics[width=1\linewidth]{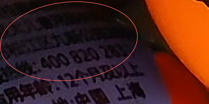}}
			\end{minipage}
			\vfill
			\begin{minipage}{0.32\linewidth}
				\centerline{\includegraphics[width=1\linewidth]{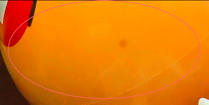}}
			\end{minipage}
			\hfill
			\begin{minipage}{0.32\linewidth}
				\centerline{\includegraphics[width=1\linewidth]{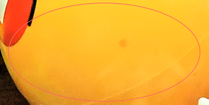}}
			\end{minipage}
			\hfill
			\begin{minipage}{0.32\linewidth}
				\centerline{\includegraphics[width=1\linewidth]{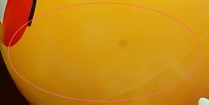}}
			\end{minipage}
			\vfill
			\begin{minipage}{0.32\linewidth}
				\centerline{\includegraphics[width=1\linewidth]{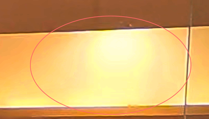}}
			\end{minipage}
			\hfill
			\begin{minipage}{0.32\linewidth}
				\centerline{\includegraphics[width=1\linewidth]{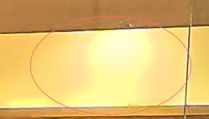}}
			\end{minipage}
			\hfill
			\begin{minipage}{0.32\linewidth}
				\centerline{\includegraphics[width=1\linewidth]{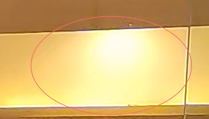}}
			\end{minipage}
			\vfill
			\begin{minipage}{0.32\linewidth}
				\scriptsize
				\centerline{Youtube~\cite{tmo_youtube}}
			\end{minipage}
			\hfill
			\begin{minipage}{0.32\linewidth}
				\scriptsize
				\centerline{UTMNet~\cite{tmo_utm}}
			\end{minipage}
			\hfill
			\begin{minipage}{0.32\linewidth}
				\scriptsize
				\centerline{GT}
			\end{minipage}
		\end{minipage}
		\hfill
		\begin{minipage}{0.48\linewidth}
			\begin{minipage}{0.32\linewidth}
				\centerline{\includegraphics[width=1\linewidth]{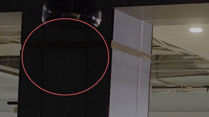}}
			\end{minipage}
			\hfill
			\begin{minipage}{0.32\linewidth}
				\centerline{\includegraphics[width=1\linewidth]{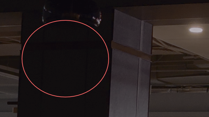}}
			\end{minipage}
			\hfill
			\begin{minipage}{0.32\linewidth}
				\centerline{\includegraphics[width=1\linewidth]{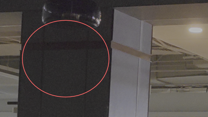}}
			\end{minipage}
			\vfill
			\begin{minipage}{0.32\linewidth}
				\centerline{\includegraphics[width=1\linewidth]{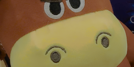}}
			\end{minipage}
			\hfill
			\begin{minipage}{0.32\linewidth}
				\centerline{\includegraphics[width=1\linewidth]{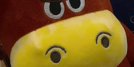}}
			\end{minipage}
			\hfill
			\begin{minipage}{0.32\linewidth}
				\centerline{\includegraphics[width=1\linewidth]{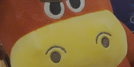}}
			\end{minipage}
			\vfill
			\begin{minipage}{0.32\linewidth}
				\centerline{\includegraphics[width=1\linewidth]{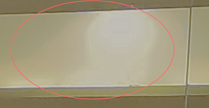}}
			\end{minipage}
			\hfill
			\begin{minipage}{0.32\linewidth}
				\centerline{\includegraphics[width=1\linewidth]{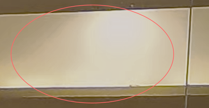}}
			\end{minipage}
			\hfill
			\begin{minipage}{0.32\linewidth}
				\centerline{\includegraphics[width=1\linewidth]{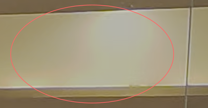}}
			\end{minipage}
			\vfill
			\begin{minipage}{0.32\linewidth}
				\scriptsize
				\centerline{Youtube~\cite{tmo_youtube}}
			\end{minipage}
			\hfill
			\begin{minipage}{0.32\linewidth}
				\scriptsize
				\centerline{UTMNet~\cite{tmo_utm}}
			\end{minipage}
			\hfill
			\begin{minipage}{0.32\linewidth}
				\scriptsize
				\centerline{GT}
			\end{minipage}
		\end{minipage}
		\vfill
		\begin{minipage}{0.48\linewidth}
			\scriptsize
			\centerline{(a) SDRTVs}
		\end{minipage}
		\hfill
		\begin{minipage}{0.48\linewidth}
			\scriptsize
			\centerline{(b) HDRTVs}
		\end{minipage}
		
	\end{center}
	\caption{(a) Drawbacks on SDRTV data synthesis of two representative TMOs~\cite{tmo_youtube,tmo_utm}. From top to bottom: information over-preservation, color bias and artifacts. (b) Reconstruction artifacts on real-world HDRTVs of the networks (HDRTVNet~\cite{chen2021new}) trained with data synthesized by these two TMOs.}
	\label{Fig_TMO_problem}
\end{figure*}

However, through detailed analysis, we observe that, because TMOs aim at preserving the information from HDRTVs as much as possible, they may inherit too much information such as extreme-light details from HDRTVs, which often do not appear in real-world SDRTVs. Such information over-preservation, as shown in Fig.~\ref{Fig_TMO_problem}(a), is the main drawback of TMOs as SDRTV data synthesis solutions. Moreover, most TMOs will also introduce color bias due to inaccurate gamut mapping and obvious artifacts such as wrong structures. Accordingly, the HDRTV reconstruction networks trained by TMO-synthesized SDRTV-HDRTV pairs are hard to generalize well to real-world cases as shown in Fig.~\ref{Fig_TMO_problem}(b).

To solve this problem, we propose an learning-based SDRTV data synthesis approach to synthesize realistic SDRTV-HDRTV pairs. Inspired by real-world degradation learning with the help of predefined degradations in super-resolution~\cite{wei2021unsupervised,maeda2020unpaired,Chen_2020_CVPR_Workshops}, we exploit the tone mapping priors in our method for both network structures and loss functions. 

In specific, we model the SDRTV data synthesis with two streams, \emph{i.e.}, a global mapping stream and a local adjustment one and use some representative tone mapping results to generate global guidance information for better HDRTV-to-SDRTV conversion. To train the network, we utilize different tone mapping results as the supervisor for regions with different light conditions, forming a novel unsupervised content loss to constraint different aspects of the synthesized SDRTVs. We also introduce an adversarial loss to emphasize the synthesized SDRTVs to be more realistic.

To validate the effectiveness of our approach, we synthesize SDRTV-HDRTV pairs using our method and use them to train several HDRTV reconstruction networks. For inference, we collect two inference datasets containing labeled SDRTVs captured by a smartphone and unlabeled SDRTVs from public datasets~\cite{kim2020jsi}. Quantitative and qualitative experimental results on these two inference datasets demonstrate that, the networks trained with our synthesized data can achieve significantly better performance than those with other data synthesis approaches.

%

\section{Related Work}

\noindent\textbf{SDRTV-to-HDRTV methods.}\quad SDRTV-to-HDRTV is a highly ill-posed problem since the complicated degradation from HDRTVs to SDRTVs. While early researches aim at restoring HDR radiance map from a low dynamic range (LDR) input, which is called inverse tone mapping~\cite{banterle2006inverse,rempel2007ldr2hdr,kovaleski2009high,10.1145/3130800.3130834,eilertsen2017hdr,lee2018deep,liu2020single,santos2020single}, they only consider HDR reconstruction at the imaging side and ignore the color gamut transform. Recently, SDRTV-to-HDRTV with deep learning techniques relying on synthesized SDRTV data becomes popular~\cite{kim2019deep,kim2020jsi,chen2021new}. In this paper, we focus on the solution of data synthesis for real-world HDRTV reconstruction.

\noindent\textbf{Tone mapping operators.}\quad TMOs aim at compressing the dynamic range of HDR sources but preserve image details as much as possible. Traditional TMOs always involve some useful tone mapping priors such as the Weber-Fechner law~\cite{drago2003adaptive} and the Retinex Theory~\cite{land1971lightness} to make either global mappings~\cite{tmo_reinhard,tmo_hable,tmo_youtube} or local mappings~\cite{tmo_chiu,tmo_kuang,tmo_liang,tmo_raman,tmo_wardHistAdj}. Recently, learning-based TMOs become popular due to their remarkable performance. They rely on ranking traditional TMOs~\cite{rana2019deep,cao2020adversarial,patel2017generative,montulet2019deep,zhang2019deep} as labels for fully supervision or unpaired datasets for adversarial learning~\cite{tmo_utm}. In this paper, we argue that TMOs have several drawbacks for realistic HDRTV data synthesis. Accordingly, we propose a learning-based method integrating tone mapping priors to solve these drawbacks.

\begin{figure}[!t]
	\begin{center}
		\begin{minipage}{0.49\linewidth}
			\centerline{\includegraphics[width=0.6\linewidth]{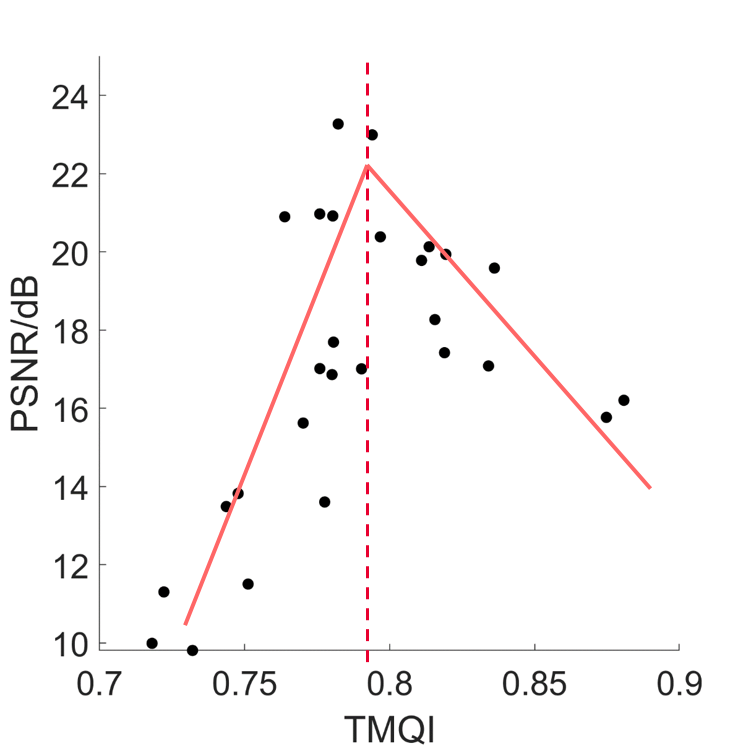}}
		\end{minipage}
		\hfill
		\begin{minipage}{0.49\linewidth}
			\centerline{\includegraphics[width=0.6\linewidth]{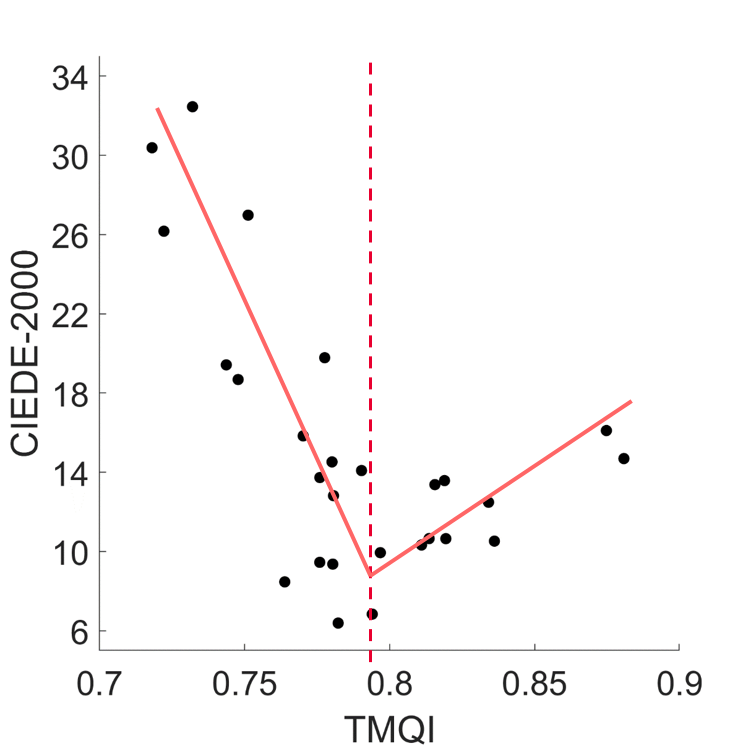}}
		\end{minipage}
		
		\vfill
		\begin{minipage}{0.49\linewidth}
			\scriptsize
			\centerline{(a)}
		\end{minipage}
		\hfill
		\begin{minipage}{0.49\linewidth}
			\scriptsize
			\centerline{(b)}
		\end{minipage}
	\end{center}
	\caption{The scatters showing statistical relationships between PSNR (a) and CIEDE-2000~\cite{sharma2005ciede2000} (b) \emph{w.r.t} TMQI~\cite{yeganeh2012objective}, respectively. We use solid lines and dotted lines to represent the trend and the turning point of trend changes, respectively. These metrics are evaluated and averaged on our collected RealHDRTV dataset.}
	\label{Fig_PSNR_TMQI}
\end{figure}

\section{Motivation}
\label{Sec_motivation}

As we all know, the core target of TMOs is to preserve as much information as possible from the HDR sources. However, the essential of the degradation from HDRTVs to SDRTVs is to lose information selectively, \emph{i.e.}, drop out details at extreme-light regions. Thus sometimes a contradiction will occur when we use TMOs to model the degradation. To get a deep-in understanding of this problem, we make an evaluation on 31 TMOs (detailed in the supplementary material) with our RealHDRTV dataset (detailed in Sec.~\ref{dataset}).

Specifically, we use TMQI~\cite{yeganeh2012objective} (higher is better), which is mostly used for the evaluations of TMOs, to evaluate the amount of information an SDRTV preserves from the corresponding HDRTV. Meanwhile, we use PSNR (higher is better) and CIEDE-2000~\cite{sharma2005ciede2000} (lower is better) to evaluate the distance between a synthesized SDRTV and the ground truth real-world one. We draw the evaluation results averaged over the RealHDRTV dataset to two scatters in Fig.~\ref{Fig_PSNR_TMQI} where each point represents a TMO.

Interestingly, we can see that, on our RealHDRTV dataset, when the TMQI of a TMO exceeds a threshold at about 0.8, the distance between synthesized and real-world data turns to increase. It indicates that the information preserved by this TMO may be too much compared with realistic SDRTVs. We can also observe such information over-preservation in Fig.~\ref{Fig_TMO_problem}(a). Such drawback may lead the trained HDRTV reconstruction networks fail to hallucinate the extreme-light details in real-world cases as shown in Fig.~\ref{Fig_TMO_problem}(b). 

Moreover, most TMOs transform the color gamut by simple transformation matrix~\cite{bt2407} or color channel rescaling~\cite{tmo_utm}, resulting obvious color bias, let alone possible artifacts such as halo, wrong structures and color banding occur for most TMOs~\cite{tmo_utm}. The data synthesized by TMOs will lead the trained reconstruction network to generate artifacts in real-world cases as shown in Fig.~\ref{Fig_TMO_problem}(b). 

Motivated by these drawbacks of TMOs on realistic SDRTV data synthesis, we propose a learning-based approach to synthesize training data for better HDRTV reconstruction in real-world cases.

\section{Learning-based SDRTV Data Synthesis}

\begin{figure*}[!t]
	\centering
	\includegraphics[width=0.88\linewidth]{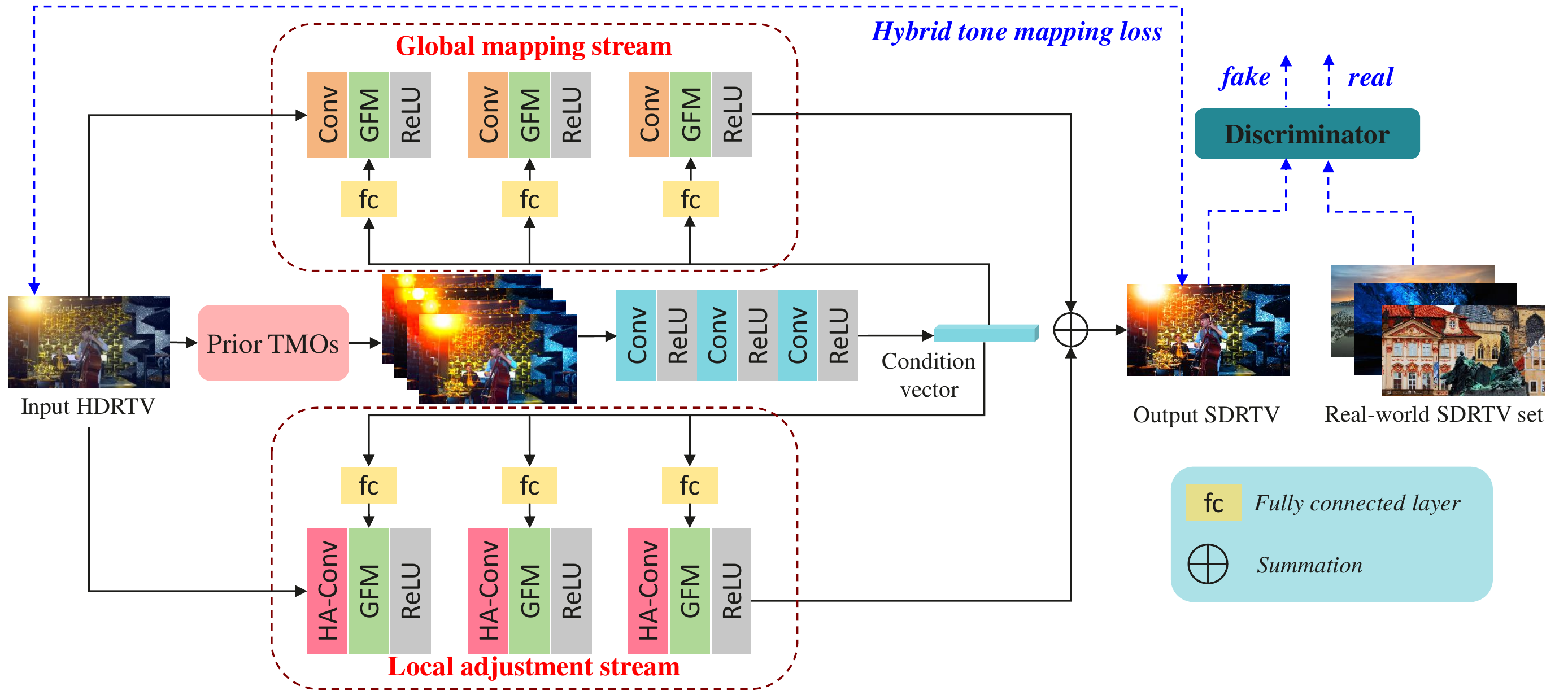}
	\caption{Our proposed SDRTV data synthesis approach. We integrate several tone mapping priors into this framework, resulting a two-stream data synthesis network conditioned by prior tone mapping results and a novel content loss function formulated by tone mapping priors.}
	\label{Fig_framework}
\end{figure*}

Fig.~\ref{Fig_framework} illustrates the framework of our data synthesis method. Inspired by learning real-world degradation with the help of predefined downsampling methods in the field of image super-resolution~\cite{maeda2020unpaired,Chen_2020_CVPR_Workshops,wei2021unsupervised}, we involve the prior knowledge for designing TMOs to our framework. Although these priors themselves cannot be used for straightforward degradation modeling, some of them can provide regional constraints or global guidance to benefit our learning. Thus, we integrate several tone mapping priors into both network structures and loss functions.

\subsection{Conditioned two-stream network}

Given an input HDRTV $H\in \mathbb{R}^{X\times Y\times 3}$ where $X$ and $Y$ denote the image size, our network $N$ aims to convert it into an SDRTV $S\in \mathbb{R}^{X\times Y\times 3}$ whose properties are similar as the real-world SDRTVs. Considering that we need both global transformations such as color gamut mapping and local adjustments such as selective detail preservation at extreme-light regions, our network $N$ includes a global mapping stream $N_{g}$ and a local adjustment stream $N_{l}$ as shown in Fig.~\ref{Fig_framework}.

The global stream $N_{g}$ is composed of three $1\times 1$ convolutions which performs similarly as global TMOs with 3DLUTs~\cite{tmo_youtube} or S-curves~\cite{tmo_reinhard,tmo_hable}. Such network has been validated effective for global color~\cite{chen2021new} and style~\cite{he2020conditional} transformations. The other stream $N_{l}$ is composed of three highlight-aware convolution blocks (HA-conv, detailed in the supplementary material), which shows superior performance on the task sensitive to extreme-light regions such as SVBRDF estimation~\cite{guo2021highlight}. For simplicity of the data synthesis network, we straightforwardly add the results of global and the local stream together to get the final synthesized SDRTVs.

Moreover, to benefit the learning, we involve the prior knowledge of existing TMOs into these two streams. For each input HDRTV $H$, we obtain a number of tone mapped versions $\{S_{i}|i=1,2,\cdots,K\}$ as the condition to guide the data synthesis. Specifically, we concatenate these condition images and feed them into a condition network $N_{c}$. The condition network is composed of three convolution layers with large kernel sizes and strides followed by a global average pooling layer. The pooling layer will output a 1D condition vector $v_{c}\in \mathbb{R}^{B\times C_{cond}}$ where $B$ and $C_{cond}$ denote the batch size and the channel number, respectively.  

Because the condition vector embeds sufficient global information of the prior tone mapping results, it is then used to modulate the main branch of the two stream network $N$. For the output feature maps $F\in \mathbb{R}^{B\times C_{feat}\times X\times Y}$ of each layer/block in the global/local stream where $C_{feat}$ denotes the channel number, we use a fully connected layer to transform $v_{c}$ to scale factors $\omega_{1}\in \mathbb{R}^{B\times C_{feat}}$ and shift factors $\omega_{2}\in \mathbb{R}^{B\times C_{feat}}$ and modulate the feature maps $F$ via global feature modulation (GFM)~\cite{he2020conditional}, which can be described as:
\begin{equation}
	F_{mod} = F * \omega_{1} + \omega_{2}.
\end{equation}
Note that we do not share the fully connected layers used for $N_{g}$ and $N_{l}$, they can provide different guidances for different transformation granularities. 


\subsection{Hybrid tone mapping prior loss}

As analyzed in Sec.~\ref{Sec_motivation}, the synthesized SDRTVs should have several aspects: globally compressed dynamic range, accurate color gamut and lost details at extreme-light regions. However, there are no paired HDRTV-SDRTV datasets and the acquisition of large-scale and high-quality datasets for training with imaging devices is also difficult. Therefore, we follow these region-aware aspects and divide the whole image into several regions according to their brightness distributions. After that, we transform the input HDRTVs with existing TMOs to get weak supervisors for different regions, forming a novel content loss function, namely hybrid tone mapping prior (HTMP) loss ($\mathcal{L}_{htmp}$).

\noindent\textbf{Region division.}\quad At the very first, we divide the input HDRTV $H$ into three regions, \emph{i.e.}, the high-, mid- and low-light regions. Specifically, we get the light radiance $L$ by linearizing $H$ with a PQ EOTF~\cite{bt2100} and segment the radiance map into three regions by two truncation points $\alpha$ and $\beta$, which are the $a$-th and $b$-th percentiles of the radiance map's histogram, respectively. The resulting region division masks are calculated as:
\begin{equation}
	M_{high} = I(L>a), M_{low}=I(L<b), M_{mid}=\textbf{1}-M_{high}-M_{low},
\end{equation}
where $I(\cdot)$ denotes the indicative function and $\textbf{1}$ is a all-one map.

\noindent\textbf{High-light loss.}\quad For the high-light regions, the output SDRTV should be saturated. Thus we use a all-one map as the supervisor at this region as:
\begin{equation}
	\mathcal{L}_{high} = \Vert M_{high}\odot (\textbf{1} - N(H))\Vert_{1},
\end{equation}
where $\odot$ means element-wise production.
Note that, although the supervisor at the high-light regions is a all-one map, due to the fact that CNNs have denoising and smoothing effects~\cite{ulyanov2018deep}, the resultant SDRTVs will become smooth here.

\noindent\textbf{Low-light loss.}\quad For the low-light regions, the output SDRTV should linearly compress the radiance due to its lower bit width. Thus we use the results of a simple TMO Linear~\cite{tmo_linear} $l_{\cdot}$ as the supervisor:
\begin{equation}
	\mathcal{L}_{low} = \Vert M_{low}\odot (l(H) - N(H))\Vert_{1}.
\end{equation}

\noindent\textbf{Mid-light loss.}\quad For the mid-light regions, we need to consider both global dynamic range compression and accurate color gamut. However, there is no proper TMO for both properties. Thus we combine two TMOs to achieve this goal. In specific, we firstly use a $\mu$-law function~\cite{kalantari2017deep} $\mu(\cdot)$ after global color gamut mapping~\cite{bt2407} $CGM(\cdot)$. Since the $\mu$-law function is a logarithm curve, which is similar to the compressive response to light in the human visual system, \emph{i.e.}, the Weber-Fechner law~\cite{drago2003adaptive}, it can provide a visually pleasant global transformation for dynamic range compression and preserve low-light details by stretching the brightness. Meanwhile, such stretching will lead to under-saturated color, so we then introduce another TMO Youtube~\cite{tmo_youtube} $y(\cdot)$, which uses 3D lookup tables predefined by Youtube tools for online film showcase. Youtube can provide vivid but sometimes over-saturated color. Moreover, due to its point-wise processing nature, Youtube will generate discontinuous textures near the high-light regions. Because the $\mu$-law function and Youtube are complementary to each other, we use an invert $\mu$-law function, \emph{i.e.}, $\mu^{-1}(\cdot)$ with the normalized linear radiance as input to generate a weighting matrix $W=\mu^{-1}(\frac{L-\beta}{\alpha-\beta})$. So the loss function at the mid-light regions can be described as:
\begin{equation}
	\mathcal{L}_{mid} = \Vert M_{mid}\odot (W\odot \mu(CGM(H)) + (\textbf{1} - W)\odot y(H)) - N(H) \Vert_{1}.
\end{equation}

Finally, we add the above three loss functions, forming our HTMP loss via $\mathcal{L}_{htmp} = \mathcal{L}_{high}+\mathcal{L}_{mid}+\mathcal{L}_{low}$. We also illustrate a flowchart of our HTMP loss for a more intuitive understanding in the supplementary material.

\subsection{Adversarial loss}
\label{Sec_adv}

With the content loss $\mathcal{L}_{htmp}$, the network has had the ability to model the region-aware properties of realistic SDRTVs. To further emphasize the synthesized SDRTVs to be more realistic, we introduce an additional adversarial loss with a discriminator following the GAN-based low-level researches~\cite{ledig2017photo}. Specifically, we collect a large real-world SDRTV dataset $\mathcal{S}$ containing $3603$ 4K SDRTVs from public datasets~\cite{kim2020jsi}. We split the dataset into train and inference subsets $\mathcal{S}_{train}$ and $\mathcal{S}_{test}$ while the latter contains $25$ SDRTVs. The dataset $\mathcal{S}$ contains SDRTVs captured in different environments and with different devices. 


During the adversarial training, we utilize the least square GAN approach~\cite{mao2017least} with a $70\times 70$ PatchGAN~\cite{isola2017image,ledig2017photo,li2016precomputed,zhu2017unpaired} and the overall loss function for the generator network $N$ is $\mathcal{L}_{N} = \mathcal{L}_{htmp}+\lambda \mathcal{L}_{adv}$, where $\lambda$ is a weighting factor. More implementation details can be found in the supplementary material.


\section{Experimental results}

\subsection{Experimental settings}
\label{dataset}

For the training of our SDRTV data synthesis network $N$, we collect a dataset $\mathcal{H}$ containing $3679$ HDRTVs (BT.2020 with PQ EOTF~\cite{bt2407}) from public datasets~\cite{kim2020jsi} as the input of network $N$. To validate the effectiveness of our the trained data synthesis network, we firstly train several HDRTV reconstruction networks using the SDRTV-HDRTV pairs synthesized by our well-trained $N$. Then we inference these networks on two real-world SDRTV datasets to see the generalization ability of trained networks.

\noindent\textbf{Datasets.}\quad With the unlabeled inference dataset $\mathcal{S}_{test}$ introduced in Sec.~\ref{Sec_adv}, we can only make visual comparisons and user study to validate the quality of reconstructed HDRTVs. In order to make full-reference evaluations, we also capture a dataset, named RealHDRTV, containing SDRTV-HDRTV pairs. Specifically, we capture 93 SDRTV-HDRTV pairs with 8K resolutions using a smartphone camera with the ``SDR" and ``HDR10" modes. To avoid possible misalignment, we use a professional steady tripod and only capture indoor or controlled static scenes. After the acquisition, we cut out regions with obvious motions ($10+$ pixels) and light condition changes, crop them into 4K image pairs and use a global 2D translation to align the cropped image pairs~\cite{chen2019camera}. Finally, we remove the pairs which are still with obvious misalignment and get $97$ 4K SDRTV-HDRTV pairs with misalignment no more than 1 pixel as our labeled inference dataset. We've release the RealHDRTV dataset in \url{https://github.com/huawei-noah/benchmark}. More details about the dataset acquisition and post-processing can be found in the supplementary material. 

\noindent\textbf{Data synthesis baselines.}\quad As for baseline SDRTV synthesis methods, we use three traditional TMOs, \emph{i.e.}, Youtube~\cite{tmo_youtube}, Hable~\cite{tmo_hable} and Raman~\cite{tmo_raman} because they are often used for film showcase in different online video platforms. We then collect other $27$ traditional TMOs (detailed in the supplementary material) and rank the $30$ TMOs using TMQI~\cite{yeganeh2012objective} and choose the best one as a new baseline named Rank following~\cite{rana2019deep,cao2020adversarial,patel2017generative,montulet2019deep}. In addition, the state-of-the-art learning-based TMO, named UTMNet~\cite{tmo_utm} is also involved here for SDRTV synthesis.

\noindent\textbf{HDRTV reconstruction networks.}\quad We use the public HDRTV dataset HDRTV1K~\cite{chen2021new} as the input of both our well-trained network $N$ and other five baselines to synthesize SDRTV-HDRTV pairs. As a result, we get $6$ datasets named after their synthesis methods to train HDRTV reconstruction networks. Specifically, we choose four state-of-the-art networks (JSI-Net~\cite{kim2020jsi}, CSRNet~\cite{he2020conditional}, SpatialA3DLUT~\cite{wang2021real}, and HDRTVNet-AGCM~\cite{chen2021new}). To compare with existing unpaired learning-based reconstruction methods, we also involve CycleGAN~\cite{zhu2017unpaired} as another reconstruction network. Note that because CycleGAN has no explicit modeling of the unique relationships between SDRTVs and HDRTVs, we do not involve it as a data synthesis baseline. The implementation details of these networks can be found in the supplementary material.

\noindent\textbf{Evaluation metrics.}\quad With the labeled dataset, \emph{i.e.}, our RealHDRTV dataset, we evaluate the reconstructed HDRTVs using several metrics for fidelity, perceptual quality and color difference. For fidelity, we use PSNR, mPSNR~\cite{banterle2017advanced}, SSIM~\cite{wang2004image}, and MS-SSIM~\cite{wang2003multiscale}. For perceptual quality, we use HDR-VDP-3~\cite{mantiuk2011hdr} and SR-SIM~\cite{zhang2012sr} because they are highly correlated to the human perceptions for HDRTVs~\cite{athar2019perceptual}. For color difference, we utilize $\triangle E_{ITP}$~\cite{bt2124} which is designed for the color gamut BT.2020. For visualization, we visualize HDRTVs without any post-processing following~\cite{chen2021new} to keep the details in extreme-light regions.

\subsection{Generalize to labeled real-world SDRTVs}
\label{Sec_Exp_label}

\noindent\textbf{Quantitative results.}\quad Quantitative results on the generalization to our RealHDRTV dataset are shown in Table~\ref{tab_sdr_to_hdr_paired}. As we can see, for each network, the version trained by paired data synthesized by our method works the best in terms of every evaluation metric and achieves significant gains over the baseline methods. Taking HDRTVNet-AGCM~\cite{chen2021new}, the state-of-the-art HDRTV reconstruction network, as an example, compared with the best-performed TMO Hable~\cite{tmo_hable}, our method gains 2.60dB, 0.014 and 6.7 in terms of PSNR, SR-SIM and $\triangle E_{ITP}$, respectively. Such results validate that, with our synthesized training data, the networks can generalize well to the real-world SDRTVs. Note that there are still small misalignment between SDRTVs and HDRTVs within this dataset, the absolute full-reference metrics will be not as high as those well-aligned ones, but the metric difference can still reflect the superiority of our method.

\begin{table*}[!t]
	\scriptsize
	\begin{center}
		\begin{tabular}{c|c|cccc|cc|c}
			\toprule[1pt]
			Network & TrainData & PSNR$\uparrow$ & mPSNR$\uparrow$ & SSIM$\uparrow$ & \makecell[c]{MS-\\SSIM}$\uparrow$ & \makecell[c]{HDR-\\VDP3}$\uparrow$ & \makecell[c]{SR-\\SIM}$\uparrow$& $\triangle E_{ITP}$$\downarrow$ \\
			\hline
			\multirow{6}*{JSI-Net~\cite{kim2020jsi}} & Raman~\cite{tmo_raman} & 18.91 & 13.23 & 0.708 & 0.719 & 3.54 & 0.736 & 74.2\\
			& Rank & 17.75 & 11.42 & 0.680 & 0.668 & 4.16 & 0.723& 81.9\\
			& UTMNet~\cite{tmo_utm} & 15.68 & 8.09 & 0.598 & 0.737 & 4.26 & 0.753 & 107.3\\
			& Youtube~\cite{tmo_youtube} & 25.47 & 18.56 & 0.842 & 0.923 & 6.32 & 0.942 & 33.6\\
			& Hable~\cite{tmo_hable} & 25.45 & 19.60 & 0.851 & 0.918 & 5.71 & 0.926 & 33.8\\
			& Ours & \textbf{27.80} & \textbf{22.92} & \textbf{0.878} & \textbf{0.933} & \textbf{6.38} & \textbf{0.943} & \textbf{27.2}\\
			\hline
			\multirow{6}*{CSRNet~\cite{he2020conditional}} & Raman~\cite{tmo_raman} & 15.16 & 9.04 & 0.628 & 0.868 & 5.16 & 0.843 & 131.3\\
			& Rank & 19.41 & 13.43 & 0.749 & 0.912 & 6.28 & 0.929 & 84.0\\
			& UTMNet~\cite{tmo_utm} & 12.37 & 5.40 & 0.433 & 0.829 & 4.63 & 0.815 & 172.2\\
			& Youtube~\cite{tmo_youtube} & 25.29 & 18.30 & 0.834 & 0.923 & 6.36 & 0.945 & 34.2\\
			& Hable~\cite{tmo_hable} & 25.34 & 19.45 & 0.847 & 0.925 & 6.35 & 0.942 & 33.8\\
			& Ours & \textbf{27.73} & \textbf{22.65} & \textbf{0.874} & \textbf{0.935} & \textbf{6.43} & \textbf{0.950} & \textbf{27.2}\\
			\hline
			\multirow{6}*{\makecell[l]{Spatial-\\ A3DLUT~\cite{wang2021real}}} & Raman~\cite{tmo_raman} & 15.35 & 10.77 & 0.726 & 0.882 & 5.61 & 0.852 & 117.4\\
			& Rank & 22.68 & 16.74 & 0.829 & 0.920 & 5.90 & 0.931 & 50.1\\
			& UTMNet~\cite{tmo_utm} & 18.55 & 13.51 & 0.805 & 0.924 & 6.04 & 0.910 & 84.2\\
			& Youtube~\cite{tmo_youtube} & 25.27 & 18.23 & 0.832 & 0.921 & 6.34 & 0.943 & 34.2\\
			& Hable~\cite{tmo_hable} & 25.48 & 19.40 & 0.846 & 0.924 & 6.35 & 0.942 & 33.5\\
			& Ours & \textbf{27.56} & \textbf{22.44} & \textbf{0.871} & \textbf{0.933} & \textbf{6.37} & \textbf{0.945} & \textbf{27.7}\\
			\hline
			\multirow{6}*{\makecell[l]{HDRTVNet-\\ AGCM~\cite{chen2021new}}} & Raman~\cite{tmo_raman} & 19.35 & 13.61 & 0.749 & 0.902 & 5.90 & 0.904 & 88.6\\
			& Rank & 19.73 & 14.06 & 0.778 & 0.917 & 6.16 & 0.936 & 77.5\\
			& UTMNet~\cite{tmo_utm} & 16.34 & 10.43 & 0.649 & 0.887 & 5.39 & 0.868 & 112.4\\
			& Youtube~\cite{tmo_youtube} & 25.26 & 18.29 & 0.833 & 0.922 & 6.36 & 0.945 & 34.1\\
			& Hable~\cite{tmo_hable} & 25.44 & 19.48 & 0.847 & 0.925 & 6.36 & 0.943 & 33.6\\
			& Ours & \textbf{28.04} & \textbf{22.82} & \textbf{0.876} & \textbf{0.938} & \textbf{6.47} & \textbf{0.957} & \textbf{26.9}\\
			\hline
			CycleGAN~\cite{zhu2017unpaired} & --- & 10.70 & 8.90 & 0.743 & 0.891 & 5.59 & 0.862 & 203.7\\			
			\bottomrule[1pt]
		\end{tabular}
	\end{center}
	\caption{Evaluation results of the HDRTV reconstruction results on the RealHDRTV dataset via various networks trained on datasets synthesized by different SDRTV data synthesis methods.}
	\label{tab_sdr_to_hdr_paired}
\end{table*}

\begin{figure*}[!t]
	\begin{center}
		
		\begin{minipage}{0.10\linewidth}
			\centering
			\scriptsize
			SDRTV
		\end{minipage}
		\begin{minipage}{0.10\linewidth}
			\centering
			\scriptsize
			Raman
		\end{minipage}
		\begin{minipage}{0.10\linewidth}
			\centering
			\scriptsize
			Rank
		\end{minipage}
		\begin{minipage}{0.10\linewidth}
			\centering
			\scriptsize
			UTMNet
		\end{minipage}
		\begin{minipage}{0.10\linewidth}
			\centering
			\scriptsize
			Youtube
		\end{minipage}
		\begin{minipage}{0.10\linewidth}
			\centering
			\scriptsize
			Hable
		\end{minipage}
		\begin{minipage}{0.10\linewidth}
			\centering
			\scriptsize
			CycleGAN
		\end{minipage}
		\begin{minipage}{0.10\linewidth}
			\centering
			\scriptsize
			Ours
		\end{minipage}
		\begin{minipage}{0.10\linewidth}
			\centering
			\scriptsize
			GT
		\end{minipage}
		
		\begin{minipage}{0.10\linewidth}
			\centerline{\includegraphics[width=1\linewidth]{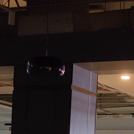}}
		\end{minipage}
		\begin{minipage}{0.10\linewidth}
			\centerline{\includegraphics[width=1\linewidth]{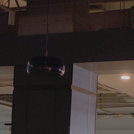}}
		\end{minipage}
		\begin{minipage}{0.10\linewidth}
			\centerline{\includegraphics[width=1\linewidth]{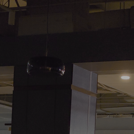}}
		\end{minipage}
		\begin{minipage}{0.10\linewidth}
			\centerline{\includegraphics[width=1\linewidth]{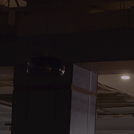}}
		\end{minipage}
		\begin{minipage}{0.10\linewidth}
			\centerline{\includegraphics[width=1\linewidth]{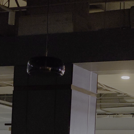}}
		\end{minipage}
		\begin{minipage}{0.10\linewidth}
			\centerline{\includegraphics[width=1\linewidth]{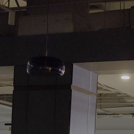}}
		\end{minipage}
		\begin{minipage}{0.10\linewidth}
			\centerline{\includegraphics[width=1\linewidth]{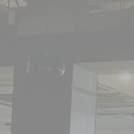}}
		\end{minipage}
		\begin{minipage}{0.10\linewidth}
			\centerline{\includegraphics[width=1\linewidth]{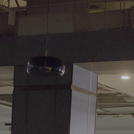}}
		\end{minipage}
		\begin{minipage}{0.10\linewidth}
			\centerline{\includegraphics[width=1\linewidth]{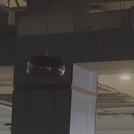}}
		\end{minipage}
		
		\begin{minipage}{0.10\linewidth}
			\centerline{\includegraphics[width=1\linewidth]{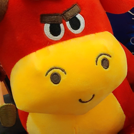}}
		\end{minipage}
		\begin{minipage}{0.10\linewidth}
			\centerline{\includegraphics[width=1\linewidth]{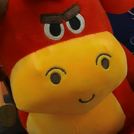}}
		\end{minipage}
		\begin{minipage}{0.10\linewidth}
			\centerline{\includegraphics[width=1\linewidth]{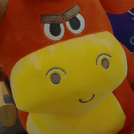}}
		\end{minipage}
		\begin{minipage}{0.10\linewidth}
			\centerline{\includegraphics[width=1\linewidth]{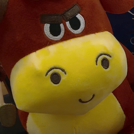}}
		\end{minipage}
		\begin{minipage}{0.10\linewidth}
			\centerline{\includegraphics[width=1\linewidth]{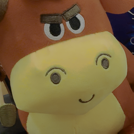}}
		\end{minipage}
		\begin{minipage}{0.10\linewidth}
			\centerline{\includegraphics[width=1\linewidth]{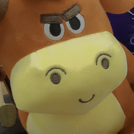}}
		\end{minipage}
		\begin{minipage}{0.10\linewidth}
			\centerline{\includegraphics[width=1\linewidth]{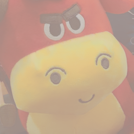}}
		\end{minipage}
		\begin{minipage}{0.10\linewidth}
			\centerline{\includegraphics[width=1\linewidth]{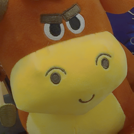}}
		\end{minipage}
		\begin{minipage}{0.10\linewidth}
			\centerline{\includegraphics[width=1\linewidth]{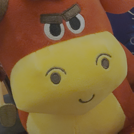}}
		\end{minipage}
		
		\begin{minipage}{0.10\linewidth}
			\centerline{\includegraphics[width=1\linewidth]{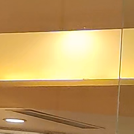}}
		\end{minipage}
		\begin{minipage}{0.10\linewidth}
			\centerline{\includegraphics[width=1\linewidth]{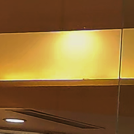}}
		\end{minipage}
		\begin{minipage}{0.10\linewidth}
			\centerline{\includegraphics[width=1\linewidth]{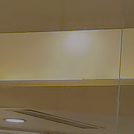}}
		\end{minipage}
		\begin{minipage}{0.10\linewidth}
			\centerline{\includegraphics[width=1\linewidth]{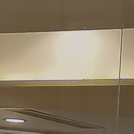}}
		\end{minipage}
		\begin{minipage}{0.10\linewidth}
			\centerline{\includegraphics[width=1\linewidth]{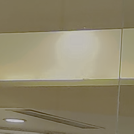}}
		\end{minipage}
		\begin{minipage}{0.10\linewidth}
			\centerline{\includegraphics[width=1\linewidth]{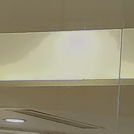}}
		\end{minipage}
		\begin{minipage}{0.10\linewidth}
			\centerline{\includegraphics[width=1\linewidth]{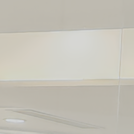}}
		\end{minipage}
		\begin{minipage}{0.10\linewidth}
			\centerline{\includegraphics[width=1\linewidth]{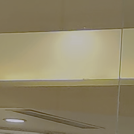}}
		\end{minipage}
		\begin{minipage}{0.10\linewidth}
			\centerline{\includegraphics[width=1\linewidth]{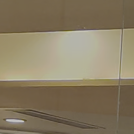}}
		\end{minipage}		
	\end{center}
	\caption{Visual comparisons on real-world SDRTV-HDRTV pairs and the HDRTVs reconstructed by HDRTVNet-AGCM~\cite{chen2021new} trained with different data synthesis methods. The images are from our RealHDRTV dataset. Zoom in the figure for a better visual experience.}
	\label{Fig_visual_labeled}
\end{figure*}

\noindent\textbf{Qualitative results.}\quad We also show some visual examples in Fig.~\ref{Fig_visual_labeled}, we can see that, with CycleGAN, the reconstructed HDRTVs suffer from severe color bias and lose details at extreme light regions, which is consistent with the results shown in Table~\ref{tab_sdr_to_hdr_paired}. Although the cycle consistency has been proved useful for style transfer~\cite{zhu2017unpaired}, the real-world HDRTV reconstruction does not work well with such constraint. In contrast, by exploiting several tone mapping priors as both constraints and guidance, our method can perform pretty well in real-world cases. While the networks trained with data synthesized by other methods show weak ability to recover the low-light region and expand the accurate color gamut, the network trained by our dataset show significant advantage over them and produce results much more close to the ground truth.

\subsection{Generalize to unlabeled real-world SDRTVs}
\label{Sec_Exp_unlabel}

We also reveal the generalization ability of the networks trained with our synthesized dataset in a more open situation, we compare the performance of three versions (Hable, Youtube, and Ours) of the network HDRTVNet-AGCM on the unlabeled inference dataset $\mathcal{S}_{test}$ collected from public datasets~\cite{kim2020jsi}. 

\begin{table}[!t]
	\scriptsize
	\centering
	\begin{tabular}{c|c|c|c|c}
		\hline
		& Hable & Youtube & Ours & \textbf{Total}\\
		\hline
		Hable & -- & 125 & 70 & \textbf{195}\\
		\hline
		Youtube & 150 & -- & 83 & \textbf{233}\\
		\hline
		Ours & 205 & 192 & -- & \textit{\textbf{397}}\\
		\hline
	\end{tabular}
	\caption{The preference matrix from the user study on the unlabeled dataset $\mathcal{S}_{test}$.}
	\label{tab_real_unlabel}
\end{table}

\begin{figure*}[!t]
	\begin{center}
		
		\begin{minipage}{0.24\linewidth}
			\centering
			\small
			SDRTV
		\end{minipage}
		\begin{minipage}{0.24\linewidth}
			\centering
			\small
			Youtube
		\end{minipage}
		\begin{minipage}{0.24\linewidth}
			\centering
			\small
			Hable
		\end{minipage}
		\begin{minipage}{0.24\linewidth}
			\centering
			\small
			Ours
		\end{minipage}
		
		\begin{minipage}{0.24\linewidth}
			\centerline{\includegraphics[width=1\linewidth]{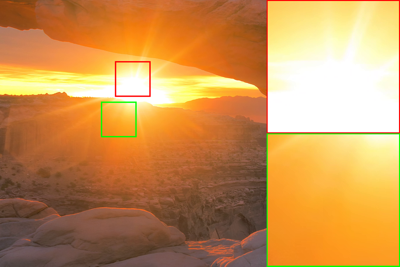}}
		\end{minipage}
		\begin{minipage}{0.24\linewidth}
			\centerline{\includegraphics[width=1\linewidth]{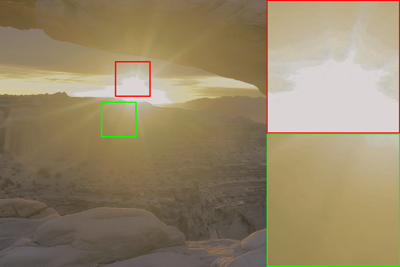}}
		\end{minipage}
		\begin{minipage}{0.24\linewidth}
			\centerline{\includegraphics[width=1\linewidth]{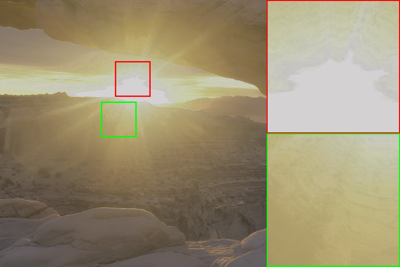}}
		\end{minipage}
		\begin{minipage}{0.24\linewidth}
			\centerline{\includegraphics[width=1\linewidth]{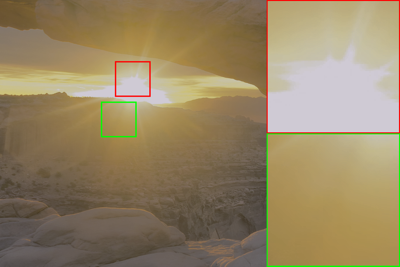}}
		\end{minipage}
		
		\begin{minipage}{0.24\linewidth}
			\centerline{\includegraphics[width=1\linewidth]{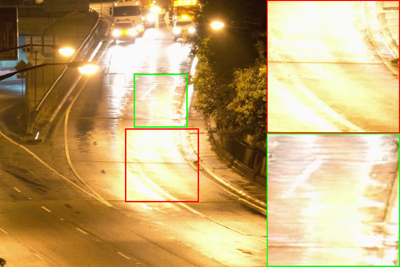}}
		\end{minipage}
		\begin{minipage}{0.24\linewidth}
			\centerline{\includegraphics[width=1\linewidth]{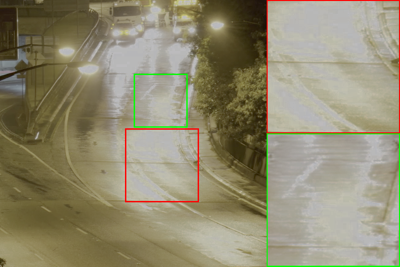}}
		\end{minipage}
		\begin{minipage}{0.24\linewidth}
			\centerline{\includegraphics[width=1\linewidth]{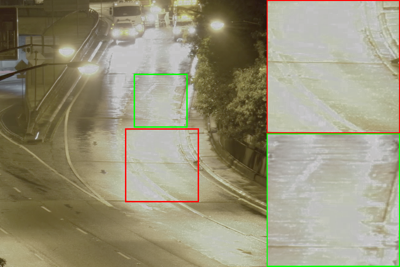}}
		\end{minipage}
		\begin{minipage}{0.24\linewidth}
			\centerline{\includegraphics[width=1\linewidth]{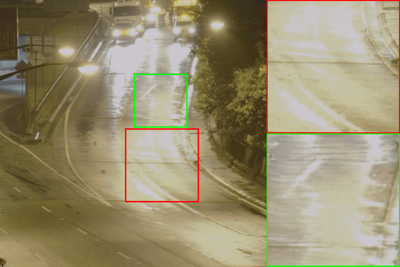}}
		\end{minipage}
		
		\begin{minipage}{0.24\linewidth}
			\centerline{\includegraphics[width=1\linewidth]{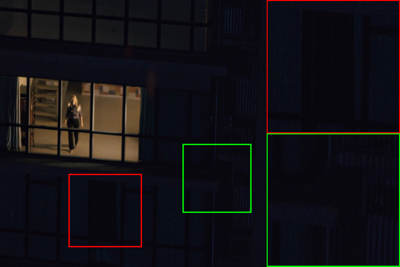}}
		\end{minipage}
		\begin{minipage}{0.24\linewidth}
			\centerline{\includegraphics[width=1\linewidth]{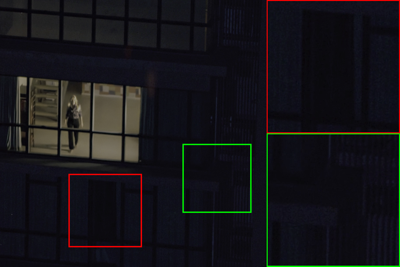}}
		\end{minipage}
		\begin{minipage}{0.24\linewidth}
			\centerline{\includegraphics[width=1\linewidth]{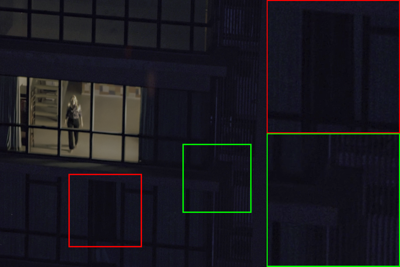}}
		\end{minipage}
		\begin{minipage}{0.24\linewidth}
			\centerline{\includegraphics[width=1\linewidth]{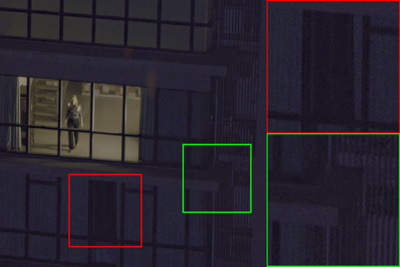}}
		\end{minipage}
		
		\begin{minipage}{0.24\linewidth}
			\centerline{\includegraphics[width=1\linewidth]{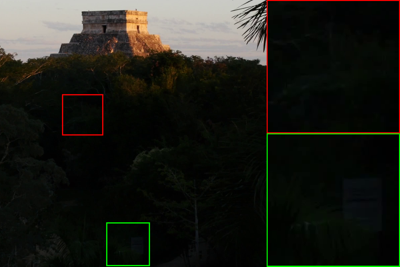}}
		\end{minipage}
		\begin{minipage}{0.24\linewidth}
			\centerline{\includegraphics[width=1\linewidth]{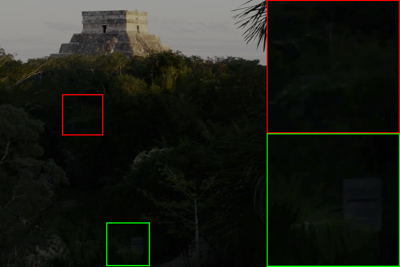}}
		\end{minipage}
		\begin{minipage}{0.24\linewidth}
			\centerline{\includegraphics[width=1\linewidth]{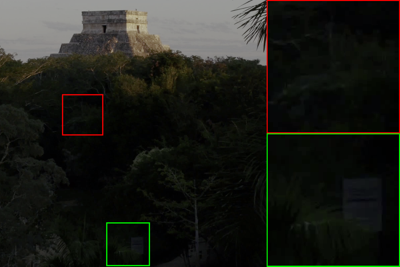}}
		\end{minipage}
		\begin{minipage}{0.24\linewidth}
			\centerline{\includegraphics[width=1\linewidth]{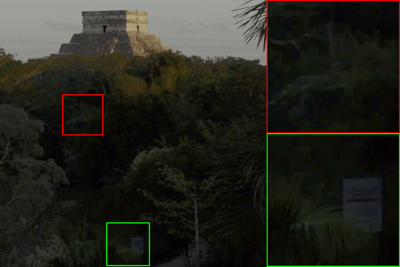}}
		\end{minipage}
		
		
	\end{center}
	\caption{Visual comparisons on the HDRTVs reconstructed by HDRTVNet-AGCM~\cite{chen2021new} trained with data synthesized by Youtube~\cite{tmo_youtube}, Hable~\cite{tmo_hable} and Ours. The input SDRTVs are from the dataset $\mathcal{S}_{test}$. Zoom in the figure for a better visual experience.}
	\label{Fig_visual_unlabeled}
\end{figure*}

\noindent\textbf{User study.}\quad We conduct a user study on the reconstructed HDRTVs with $11$ professional photographers for subjective evaluation. Each participant is asked to make pairwise comparisons on $3$ results of each image displayed on an HDR-TV in a darkroom. The detailed settings and device parameters can be found in the supplementary material. We show the preference matrix in Table~\ref{tab_real_unlabel}. We can see that, when comparing our method with the best-performed TMOs, \emph{i.e.}, Hable and Youtube, 74.5\% and 69.8\% of users prefer our results, respectively.

\noindent\textbf{Qualitative results.}\quad We also show some examples for visual comparison in Fig.~\ref{Fig_visual_unlabeled}. We can find that while the networks trained by Youtube's and Hable's data has less awareness of high-light (the top two) and low-light (the bottom two) regions, the network trained by our data can enrich the details as well as preserve continuous structures.

To sum up, while the training datasets for both our data synthesis network and the HDRTV reconstruction networks have no overlap with our RealHDRTV and $\mathcal{S}_{test}$ datasets, the networks trained by our data show notable performance gains in both numerical and visual comparisons as well as the user study. It indicates that, our approach can serve as a better solution for paired SDRTV-HDRTV synthesis towards real-world HDRTV reconstruction.

\subsection{The quality of synthesized SDRTVs}

We also evaluate the quality of synthesized SDRTVs. Specifically, we feed the HDRTVs in our RealHDRTV dataset into our well-trained data synthesis network and evaluate the distance and difference between our synthesized SDRTVs and the real-world ones. We evaluate the distances in terms of fidelity metrics PSNR and SSIM~\cite{wang2004image} and color difference for color gamut BT.709, \emph{i.e.}, CIEDE-2000~\cite{sharma2005ciede2000}. Following the experiment in Sec.~\ref{Sec_motivation}, we also calculate TMQI~\cite{yeganeh2012objective} to evaluate the ability of information preservation from HDRTVs. We involve several representative TMOs for the comparison as shown in Table~\ref{tab_hdr_to_sdr_res_one_column2}.

\begin{table}[t]
	\scriptsize
	\centering
	\begin{tabular}{c|ccc|c}
		
		\toprule[1pt]
		& PSNR$\uparrow$ & SSIM $\uparrow$ & CIEDE $\downarrow$ & TMQI $\uparrow$\\
		\hline
		Clip & 13.82 & 0.719 & 18.68 & 0.7477\\
		Linear~\cite{tmo_linear} & 16.46 & 0.758 & 15.15 & 0.7353\\
		Reinhard~\cite{tmo_reinhard} & 19.94 & 0.776 & 10.65 & 0.8194\\
		Raman~\cite{tmo_raman} & 20.97 & 0.627 & 9.52 & 0.7759\\
		Kuang~\cite{tmo_kuang} & 20.92 & 0.717 & 9.35 & 0.7804\\
		Youtube~\cite{tmo_youtube} & 22.99 & 0.824 & 6.83 & 0.7940\\
		Hable~\cite{tmo_hable} & 23.27 & 0.840 & 6.38 & 0.7822\\
		Liang~\cite{tmo_liang} & 16.21 & 0.676 & 14.81 & 0.8807\\
		Rank~\cite{rana2019deep} & 16.57 & 0.692 & 14.32 & \textbf{0.8850}\\
		UTMNet~\cite{tmo_utm} & 15.77 & 0.681 & 16.14 & 0.8747\\
		\hline
		Ours & \textbf{24.54} & \textbf{0.844} & \textbf{5.80} & 0.7988\\
		\bottomrule[1pt]
	\end{tabular}
	\caption{Evaluation metrics on fidelity and color difference between the SDRTVs synthesized by several methods and the ground truth ones on our RealHDRTV dataset.}
	\label{tab_hdr_to_sdr_res_one_column2}
\end{table}

We can observe that, although the state-of-the-art TMOs like Liang~\cite{tmo_liang} and UTMNet~\cite{tmo_utm} have significantly high TMQI values, the SDRTVs generated by them are far away from the ground truth SDRTVs. On the contrary, the SDRTVs generated by our method shows much better fidelity and color accuracy by 1.27dB gain of PSNR and 0.58 drop of CIEDE-2000 compared with the best performed TMO Hable~\cite{tmo_hable}. Such results are consistent with what we observe in Fig.~\ref{Fig_PSNR_TMQI} and interestingly, we find that our average TMQI value is pretty close to the turning point in the scatters, \emph{i.e.}, about $0.8$ for this dataset. It reveals our success on avoiding information over-preservation.

\begin{figure*}[!t]
	\begin{center}
		
		\begin{minipage}{0.092\linewidth}
			\centering
			\scriptsize
			HDRTV
		\end{minipage}
		\begin{minipage}{0.092\linewidth}
			\centering
			\scriptsize
			Clip
		\end{minipage}
		\begin{minipage}{0.092\linewidth}
			\centering
			\scriptsize
			Linear
		\end{minipage}
		\begin{minipage}{0.092\linewidth}
			\centering
			\scriptsize
			Reinhard
		\end{minipage}
		\begin{minipage}{0.092\linewidth}
			\centering
			\scriptsize
			Youtube
		\end{minipage}
		\begin{minipage}{0.092\linewidth}
			\centering
			\scriptsize
			Rank
		\end{minipage}
		\begin{minipage}{0.092\linewidth}
			\centering
			\scriptsize
			UTMNet
		\end{minipage}
		\begin{minipage}{0.092\linewidth}
			\centering
			\scriptsize
			Hable
		\end{minipage}
		\begin{minipage}{0.092\linewidth}
			\centering
			\scriptsize
			Ours
		\end{minipage}
		\begin{minipage}{0.092\linewidth}
			\centering
			\scriptsize
			GT
		\end{minipage}
		
		\begin{minipage}{0.092\linewidth}
			\centerline{\includegraphics[width=1\linewidth]{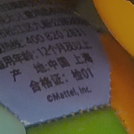}}
		\end{minipage}
		\begin{minipage}{0.092\linewidth}
			\centerline{\includegraphics[width=1\linewidth]{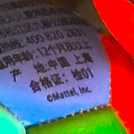}}
		\end{minipage}
		\begin{minipage}{0.092\linewidth}
			\centerline{\includegraphics[width=1\linewidth]{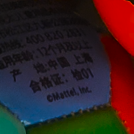}}
		\end{minipage}
		\begin{minipage}{0.092\linewidth}
			\centerline{\includegraphics[width=1\linewidth]{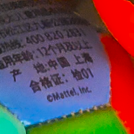}}
		\end{minipage}
		\begin{minipage}{0.092\linewidth}
			\centerline{\includegraphics[width=1\linewidth]{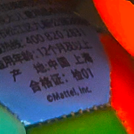}}
		\end{minipage}
		\begin{minipage}{0.092\linewidth}
			\centerline{\includegraphics[width=1\linewidth]{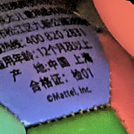}}
		\end{minipage}
		\begin{minipage}{0.092\linewidth}
			\centerline{\includegraphics[width=1\linewidth]{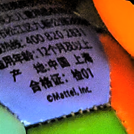}}
		\end{minipage}
		\begin{minipage}{0.092\linewidth}
			\centerline{\includegraphics[width=1\linewidth]{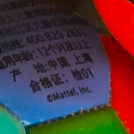}}
		\end{minipage}
		\begin{minipage}{0.092\linewidth}
			\centerline{\includegraphics[width=1\linewidth]{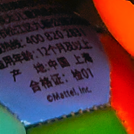}}
		\end{minipage}
		\begin{minipage}{0.092\linewidth}
			\centerline{\includegraphics[width=1\linewidth]{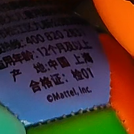}}
		\end{minipage}

		\begin{minipage}{0.092\linewidth}
			\centerline{\includegraphics[width=1\linewidth]{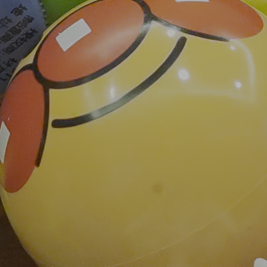}}
		\end{minipage}
		\begin{minipage}{0.092\linewidth}
			\centerline{\includegraphics[width=1\linewidth]{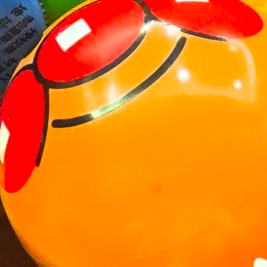}}
		\end{minipage}
		\begin{minipage}{0.092\linewidth}
			\centerline{\includegraphics[width=1\linewidth]{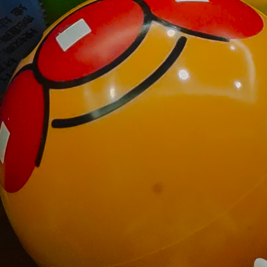}}
		\end{minipage}
		\begin{minipage}{0.092\linewidth}
			\centerline{\includegraphics[width=1\linewidth]{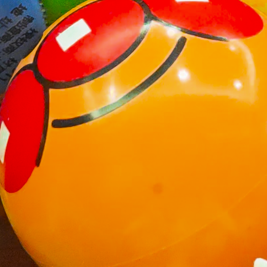}}
		\end{minipage}
		\begin{minipage}{0.092\linewidth}
			\centerline{\includegraphics[width=1\linewidth]{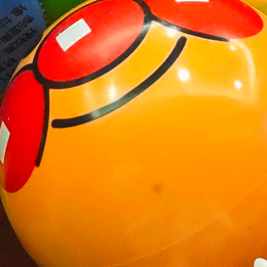}}
		\end{minipage}
		\begin{minipage}{0.092\linewidth}
			\centerline{\includegraphics[width=1\linewidth]{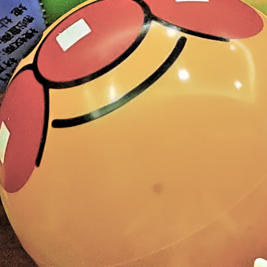}}
		\end{minipage}
		\begin{minipage}{0.092\linewidth}
			\centerline{\includegraphics[width=1\linewidth]{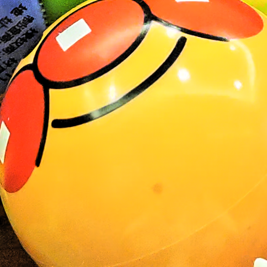}}
		\end{minipage}
		\begin{minipage}{0.092\linewidth}
			\centerline{\includegraphics[width=1\linewidth]{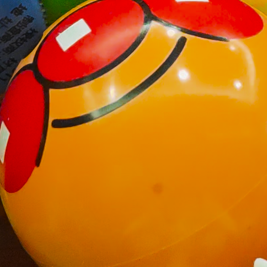}}
		\end{minipage}
		\begin{minipage}{0.092\linewidth}
			\centerline{\includegraphics[width=1\linewidth]{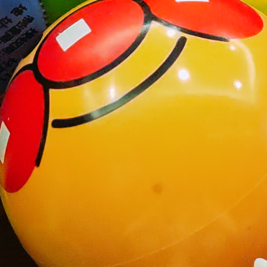}}
		\end{minipage}
		\begin{minipage}{0.092\linewidth}
			\centerline{\includegraphics[width=1\linewidth]{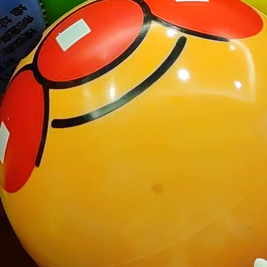}}
		\end{minipage}
		
		\begin{minipage}{0.092\linewidth}
			\centerline{\includegraphics[width=1\linewidth]{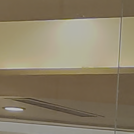}}
		\end{minipage}
		\begin{minipage}{0.092\linewidth}
			\centerline{\includegraphics[width=1\linewidth]{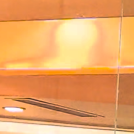}}
		\end{minipage}
		\begin{minipage}{0.092\linewidth}
			\centerline{\includegraphics[width=1\linewidth]{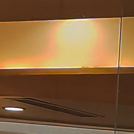}}
		\end{minipage}
		\begin{minipage}{0.092\linewidth}
			\centerline{\includegraphics[width=1\linewidth]{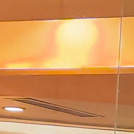}}
		\end{minipage}
		\begin{minipage}{0.092\linewidth}
			\centerline{\includegraphics[width=1\linewidth]{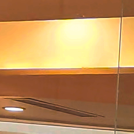}}
		\end{minipage}
		\begin{minipage}{0.092\linewidth}
			\centerline{\includegraphics[width=1\linewidth]{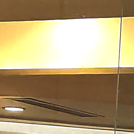}}
		\end{minipage}
		\begin{minipage}{0.092\linewidth}
			\centerline{\includegraphics[width=1\linewidth]{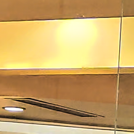}}
		\end{minipage}
		\begin{minipage}{0.092\linewidth}
			\centerline{\includegraphics[width=1\linewidth]{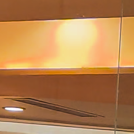}}
		\end{minipage}
		\begin{minipage}{0.092\linewidth}
			\centerline{\includegraphics[width=1\linewidth]{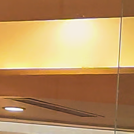}}
		\end{minipage}
		\begin{minipage}{0.092\linewidth}
			\centerline{\includegraphics[width=1\linewidth]{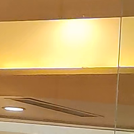}}
		\end{minipage}		
		
	\end{center}
	\caption{Visual comparisons on SDRTVs synthesized by different representative synthesis methods together with the input HDRTVs and ground truth SDRTVs. The images are from our RealHDRTV dataset. Zoom in the figure for a better visual experience.}
	\label{Fig_visual_sdrtv}
\end{figure*}

We also show some visual examples in Fig.~\ref{Fig_visual_sdrtv}. We can see that, compared with the ground truth SDRTVs, the information over-preservation (\emph{e.g.}, Clip and Rank for the top example), color bias (\emph{e.g.}, Hable and UTMNet for the middle example) and artifacts (\emph{e.g.}, Reinhard and Linear in the bottom example) are very obvious. Meanwhile, our method can selectively preserve the information from the HDRTVs, transform color gamut accurately and avoid the artifacts. 

\subsection{Ablation}
\label{ablation}

With the evaluations on synthesized SDRTVs, we'd like to show some ablation studies about the network structures and loss functions, particularly the effects on the tone mapping priors we utilize in our framework. We compare the values of PSNR and CIEDE-2000~\cite{sharma2005ciede2000} calculated on the synthesized SDRTVs by different variants in Table~\ref{tab_ablation} and show visual comparisons in the supplementary material.

\noindent\textbf{Network design.}\quad We conduct several experiments to validate the effectiveness of tone mapping priors used for network designs. Specifically, we remove the condition network or use the input HDRTV itself to replace the condition tone mapping results to keep the parameter numbers the same. We can see in Table~\ref{tab_ablation} that the condition network as well as the condition tone mapping results make very important contributions to more accurate real-world data synthesis. As Fig.~\ref{Fig_visual_sdrtv} shows that, the condition TMOs, \emph{i.e.}, Clip, Linear, Reinhard and Youtube we use here shows different performance advantages at different regions. For example, Linear performs very well at losing low-light details. Meanwhile, our method apparently take merits of these conditions, which validates the importance of them again. In addition, the ablation results on only the global or local stream validate the effectiveness of combining them, the visual results in the supplementary material also validate the advantages of these two streams on global and local mappings, respectively.

\noindent\textbf{Loss function.}\quad As we can see in the table, if we only use $\mathcal{L}_{adv}$ to train the network, the network will synthesize SDRTVs far away from the real-world ones due to the lack of content and structure constraints. However, it does not mean that $\mathcal{L}_{adv}$ is useless, we can see that with the help of $\mathcal{L}_{adv}$, the network with only $\mathcal{L}_{htmp}$ achieves a notable performance gain. In addition, to show the impacts on the involved TMOs for $\mathcal{L}_{htmp}$, we use simple $L_{1}$ loss function between the tone mapping results of each TMO as the content loss to replace $\mathcal{L}_{htmp}$. As we can see in the table, with either TMO as the supervisor, the network performance will be inferior than our $\mathcal{L}_{htmp}$. Such results validate the effectiveness of our region-aware content loss. 


\begin{table}[!t]
	\scriptsize
	\centering
	\begin{tabular}{cc|ccc|c|c}
		\hline
		\multicolumn{2}{c|}{Loss} & \multicolumn{3}{c|}{Network} & \multirow{2}*{PSNR} & \multirow{2}*{CIEDE}\\ \cline{1-2} \cline{3-5}
		$\mathcal{L}_{htmp}$ & $\mathcal{L}_{adv}$ & $N_{c}$ & $N_{l}$ & $N_{g}$ & & \\
		\hline
		$\checkmark$ & $\checkmark$ & $\checkmark$ & $\checkmark$ & $\checkmark$ & \textbf{24.54} & \textbf{5.80} \\
		\hline
		$\times$ & $\checkmark$ & $\checkmark$ & $\checkmark$ & $\checkmark$ & 11.74 & 27.01\\
		$\checkmark$ & $\times$ & $\checkmark$ & $\checkmark$ & $\checkmark$ & 24.24 & 6.00\\
		\hline
		\textit{S-Linear} & $\checkmark$ & $\checkmark$ & $\checkmark$ & $\checkmark$ & 16.35 & 15.33 \\
		\textit{S-$\mu$-law} & $\checkmark$ & $\checkmark$ & $\checkmark$ & $\checkmark$ & 18.18 & 12.42 \\
		\textit{S-Youtube} & $\checkmark$ & $\checkmark$ & $\checkmark$ & $\checkmark$ & 23.32 & 6.64 \\
		\hline
		$\checkmark$ & $\checkmark$ & $\times$ & $\checkmark$ & $\checkmark$ & 24.08 & 6.03\\
		$\checkmark$ & $\checkmark$ & \textit{Self} & $\checkmark$ & $\checkmark$ & 24.15 & 5.97\\
		$\checkmark$ & $\checkmark$ & $\checkmark$ & $\times$ & $\checkmark$ & 24.33 & 5.99 \\
		$\checkmark$ & $\checkmark$ & $\checkmark$ & $\checkmark$ & $\times$ & 24.44 & 5.83 \\
		\hline	
	\end{tabular}
	\caption{Ablation study on the RealHDRTV dataset.}
	\label{tab_ablation}
\end{table}

\section{Conclusion}

In this paper, we propose a data synthesis approach to synthesize realistic SDRTV-HDRTV pairs for the training of HDRTV reconstruction networks to benefit their generalization ability on real-world cases. Through statistical and visual analysis, we observe that, existing TMOs suffer from several drawbacks on the modeling of HDRTV-to-SDRTV including information over-preservation, color bias and artifacts. To solve this problem, we propose a learning-based SDRTV data synthesis to learn the aspects of real-world SDRTVs. We integrate several tone mapping priors into both network structures and loss functions to benefit the learning. Experimental results on our collected labeled and unlabeled datasets validate that, the HDRTV reconstruction networks trained by our synthesized dataset can generalize significantly better than other methods. In addition, we believe that integrating degradation priors into degradation learning framework may also be promoted to benefit other low-level vision tasks. In the future, we also want to use our framework on MindSpore\footnote{\url{https://www.mindspore.cn/}}, which is a new deep learning computing framework.

\section*{Acknowledgments}

We acknowledge funding from National Key R\&D Program of China under
Grant 2017YFA0700800, and National Natural Science Foundation of China under Grants 62131003 and 62021001.


\clearpage
%
%
\bibliographystyle{splncs04}
\bibliography{egbib}
\end{document}